\documentclass{article}

\usepackage{arxiv}

\usepackage[utf8]{inputenc} 
\usepackage[T1]{fontenc}    
\usepackage{hyperref}       
\usepackage{url}            
\usepackage{booktabs}       
\usepackage{amsfonts}       
\usepackage{nicefrac}       
\usepackage{microtype}      
\usepackage{lipsum}		
\usepackage{graphicx}
\usepackage[square,numbers]{natbib}
\usepackage{doi}
\usepackage{amsmath}

\title{ predicting molecule-target interaction by learning biomedical network  and molecule representations}

\author{ \href{https://orcid.org/0000-0002-9157-2199}{\includegraphics[scale=0.06]{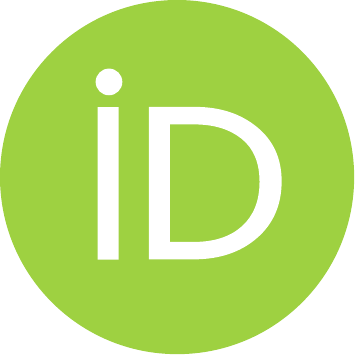}\hspace{1mm}Jinjiang Guo*}, {\hspace{2mm}Jie Li}\\
	\emph{Data Science group}\\
	Global Health Drug Discovery Institute, Beijing, China \\
	\texttt{jinjiang.guo@ghddi.org}  and \texttt{jie.li@ghddi.org}  \\
}

\date{}

\renewcommand{\headeright}{}
\renewcommand{\undertitle}{}

\hypersetup{
pdftitle={MTINeT},
pdfsubject={cs.AI},
pdfauthor={Jinjiang GUO, Jie Li},
pdfkeywords={Graph Neural Networks, Biological Network,  Link Prediction},
}

\begin{document}
\maketitle

\begin{abstract}
The study of molecule-target interaction  is quite important for drug discovery in terms of target identification, hit identification, pathway study, drug-drug interaction, etc. Most existing methodologies utilize either  biomedical network  information or molecule structural features to predict potential interaction link. However, the biomedical network information based methods  usually suffer from cold start problem, while structure based methods often give limited performance due to the structure/interaction assumption and data quality.  To address these issues, we propose a pseudo-siamese Graph Neural Network method, namely MTINet+, which learns both biomedical network topological  and molecule structural/chemical information  as  representations to predict potential interaction of given molecule and target pair. In MTINet+, 1-hop subgraphs of given molecule and target pair are extracted from known interaction of biomedical network as topological information, meanwhile the molecule structural and chemical  attributes are processed as molecule  information. MTINet+ learns these two types of information as embedding features for predicting the pair link.  In the experiments of different molecule-target interaction tasks, MTINet+ significantly outperforms over the state-of-the-art baselines. In addition, in our designed  network sparsity experiments , MTINet+ shows strong robustness against different sparse biomedical networks.
\end{abstract}


\section{Introduction}
In domains of medicine  and pharmacology, the biomedical network \cite{barabasi2011network, hopkins2008network} is widely studied in terms of drug-target interaction, drug-drug interaction, protein-protein interaction, etc., which help researchers to  better understand the disease  interactome and drug action mechanisms. Such studies \cite{campillos2008drug, morris2009autodock4, wang2014drug, zhou2020artificial, zhang2022drugai, huttlin2017architecture, li2016network} have great potency in drug repurposing, target identification, biomarker discovery, omics analysis, and drug side effects, etc.  Meanwhile, in  the studies of these biomedical networks, \emph{link prediction} \cite{yue2020graph, ezzat2019computational} is one of the important network-based computing and modelling approaches for analysing biomedicine relationships. To our knowledge, researchers usually refer to two kinds of recommendation algorithms \cite{luo2017network}, i.e., content-based recommendation methods \cite{pazzani2007content} and collaborative filtering models \cite{goldberg1992using},  as \emph{link prediction} methods. Content-based recommendation methods consider the attributes of  molecule (compound, peptide, protein, etc.) and target (compound, peptide, protein, etc.) as side information, find optimal projections between molecule and target, and to predict the molecule potential action  on the target, i.e., interactions between molecule and target. Collaborative filtering models utilize collective molecule-target links to predict potential interactions in the bipartite molecule-target network. However, since collaborative filtering models largely rely on the known molecule-target relations and consider no side information, such models often suffer from \emph{cold start} problem, which means, when a new molecule is given, the models can hardly predict its potential interacted targets.

\begin{figure}
	\centering
	\includegraphics[scale=.25]{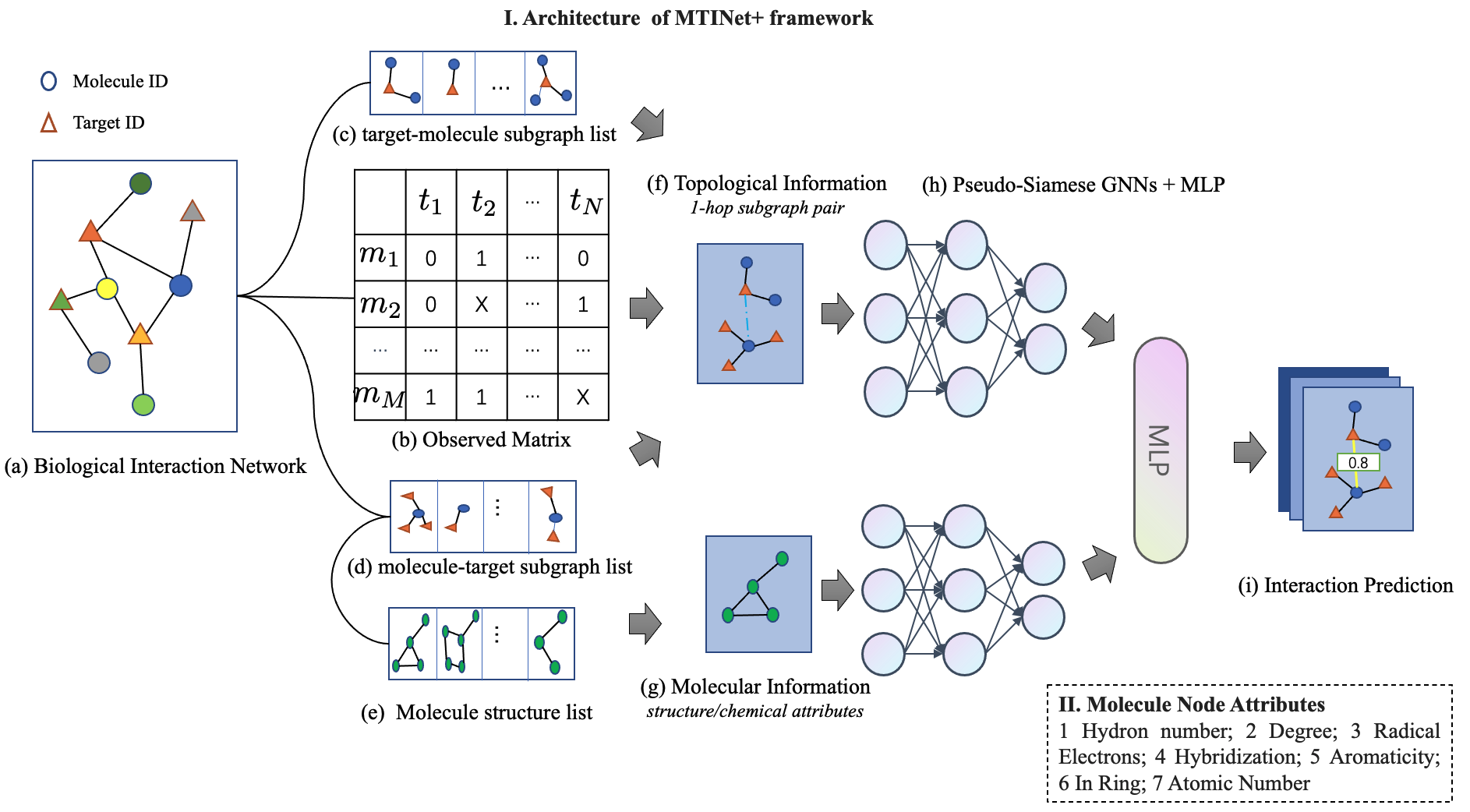}
	\caption{ I. Illustration of MTINet+:  a pseudo-siamese graph neural network framework. (a) Heterogeneous biomedical network is firstly converted into (b) observed interaction matrix  between molecules and targets. Then, the extracted 1-hop subgraphs ((d) $G^M_{m \to t}$ and (c) $G^N_{t \to m}$) around each molecule and target are paired as (f)  one  graph $\hat{G}_{m,t}^{M \times N}$ , which  is used as topological information, meanwhile each molecule SMILES are converted to (g) a graph $S_m^M$ integrated with chemical attributes, which is used as molecule information. Then, $\hat{G}_{m,t}^{M \times N}$  and $S_m^M$ are respectively fed to (h) two different GNN branches  for learning biomedical network and molecule representations.  Finally, these two kinds of representations are concatenated and infused to  a MLP functional head for learning to  predict (i) link probabilities between new input molecule and target pairs. II. Molecule node attributes. }
	\label{fig:framework}
\end{figure}

To address these issues, we propose a pseudo-siamese Graph Neural Network (GNN) method, namely Molecule-Target Interaction Network plus (MTINet+).  In this work, we convert collective molecule-target links from biomedical network into partially observed matrix, and  address  the \emph{Matrix Completion} task with link prediction on graphs.  Molecule-target interactions, such as Drug-Target Interaction (DTI), Compound-Virus Inhibition (CVI), etc., can be  described as bipartite molecule-target network  where edges denote observed links and nodes represent either molecule or target IDs. In our proposed MTINet+ framework,  1-hop subgraph is extracted around each node (molecule or target) in the bipartite molecule-target network. Such 1-hop subgraphs contain \emph{local topological pattern} which is deterministic for inferring missing links, and independent on  molecule and target side information as well \cite{zhang2019inductive}.  Exhaustively around each molecule and target node pair,  1-hop subgraphs are extracted for one  MTINet+ GNN branch to learn  the topological  representations.  Meanwhile, MTINet+ converts molecule line notation sequence (i.e., SMILES) into 2D graph formula, in which each node incorporates different chemical attributes (shown in Figure  \ref{fig:framework}(II)).  The converted molecule 2D graph is input to the other  MTINet+ GNN branch to learn the  molecule representations. Then, two kinds  of  representations are hybridized and learnt by Multi-Layer Perceptron (MLP) classifier (or regressor) for predicting potential links. Figure \ref{fig:framework} (I) illustrates the framework of our method.

The main contributions of our work are:
\begin{itemize}
	\item We developed a novel end-to-end  pseudo-siamese GNN method for predicting potential/missing links in biomedical network.
	\item MTINet+ tactfully treats both biomedical network topological information and molecule structural/chemical information as unified graph information, and learns the hybrid graph information as embedding features for link prediction.
	\item MTINet+ naturally solves the cold start problem and shows strong robustness  against different sparse biomedical networks.
	\item We curated a 56 species antiviral compound-phenotype network for researchers to fight against the COVID-19 pandemic.
\end{itemize}

This paper is organized as follows:  Section \ref{sec:relatedwork} introduces the related work. Section \ref{sec:gplpintro} gives detailed description of our MTINet+ framework. Methods comparison results on different biomedical networks are shown in Section \ref{sec:data&exp}.    Robustness analysis of MTINet+ and future work  are discussed in Section \ref{sec:dis&future}.

\section{Related work}
\label{sec:relatedwork}
In this section, we briefly introduce theories of graph neural networks (GNN) and matrix completion (MC) methods used in the studies of complex system and chemical/biological structures, such as  quantitative structure-activity relationships (QSAR),  recommendation systems and biomedical networks. Also, we give a discussion on motivations of using GNN methods to fulfil MC tasks. 

\subsection{Matrix Completion}
\label{sec:mc}
Matrix Completion (MC) is a kind of task formulation \cite{candes2009exact} commonly used  in recommender systems, which converts  heterogeneous network datasets into observed matrix $\mathcal{M}$, and aims to fill the missing entries of $\mathcal{M}$. In the matrix $\mathcal{M}$,  the rows and columns  represent  users (molecules) and items (targets) respectively, each element in $\mathcal{M}$ denotes the interaction between corresponding  (molecule, target) pair coordinate. To fulfil the MC tasks, well-known Matrix Factorization (MF) methods decompose the low-rank $\mathcal{M}$ into the product of two lower dimensionality of rectangular matrices \cite{agarwal2009regression, jannach2013recommenders}. Later on, Singular Value Decomposition (SVD) is adopted in MF families \cite{kluver2014evaluating, bi2017group, pu2013understanding}, which decompose $\mathcal{M}$ into the inner product of three matrices $X$, $Y$ and $P$ (i.e., $\mathcal{M} = XPY^T$ ), where $X$ and $Y$ contain  the latent features of molecules and targets respectively, $P$ is the association matrix. The idea is to find the projections from $X$ to $Y$ by determining the optimal $P$ from $\mathcal{M}$. SVD based MF methods can integrate  latent features of molecules and targets, and therefore partially solve the \emph{cold-start} problem \cite{kluver2014evaluating}. 

Also, SVD basd MF methods successfully solved many biomedical problems, such as Drug-Target Interaction (DTI) prediction \cite{luo2017network}, Gene-Disease Association prediction.\cite{natarajan2014inductive}, etc. In the work \cite{luo2017network}, the low-dimensional  latent features of drug (molecule) and target is obtained by integrating heterogeneous data, such like drug-drug interaction, drug-side-effect, drug-disease association, drug  similarities for molecule, and protein-protein interaction, protein-disease association, protein similarities for target. The integrated features describe topological properties for each molecule and target respectively. Similarly, for predicting Gene-Disease Association \cite{natarajan2014inductive}, microarray measurements of gene expression, gene-phenotype associations of other species  and   HumanHet \cite{lee2011prioritizing} features (incorporating mRNA expression, protein-protein interactions, protein complex data, and comparative genomics) serve as latent gene (molecule) features, while disease similarities from MimMiner \cite{van2006text} and features collected from OMIM diseases \cite{hamosh2005online} are used as latent disease (target) features. As we can see, the performance of MF methods is highly dependent upon the integrated  latent features as representations  of molecules and targets. Usually, the feature construction procedure  are  manually designed, and separate from the optimization of association matrix $P$ in  MF procedure as well.

\subsection{Spatial based Graph Neural Networks}
\label{sec:gnn}
Recently, GNN based methods  \cite{wu2018moleculenet, liu2019chemi, hamilton2017inductive, xu2018powerful, morris2019weisfeiler, ranjan2020asap} have demonstrated breakthroughs in many tasks regrading to network/structure datasets \cite{wu2020graph}, such like the studies of quantitative structure-activity relationships (QSAR), knowledge graph, physical systems \cite{battaglia2016interaction, sanchez2018graph}, etc.  And the GNN methods have outperformed  against traditional machine learning methods such as random forest (RF) \cite{pal2005random} and support vector machines (SVM) \cite{noble2006support}. There are mainly three reasons to use GNN methods in network datasets: (1) Most complex systems datasets (biomedical network data, etc.) are in the form of graph structure; (2) Molecule (compound, peptide, protein, etc.) structure can be described as graph structure as well; (3) Most importantly, GNN methods are featured at processing topological connections among nodes, and learning graph representations \cite{wu2020graph}, and therefore they can treat both biomedical network data and molecule data as unified graph information, and learn the hybrid graph information as embedding features for link prediction task.

Existing GNN methods can be categorized into two types: spectral based methods and spatial based methods \cite{wu2020graph}.  For spectral based GNNs,  graphs are projected into Fourier domain where the convolution operation is conducted by computing  Laplacian eigendecomposition \cite{defferrard2016convolutional, kipf2016semi}. Due to the high computational complexity of Laplacian eigendecomposition, the Chebyshev polynomials are adopted as an approximation \cite{defferrard2016convolutional,kipf2016semi}. Spatial based GNN methods imitate convolutional neural networks (CNN) by aggregating  and updating neighborhood message from central node \cite{hamilton2017inductive, xu2018powerful, hu2019strategies, morris2019weisfeiler, ranjan2020asap, leng2021enhance}, and construct the whole graph representation through read-out function. General operations of spatial based GNN can be expressed as below: 
\begin{equation}
\label{eq:gnn}
h_v^k = \mathbf{C}^k(h_v^{k-1}, \mathbf{A}^k(\{ (h_v^{k-1}, h_u^{k-1}, e_{uv}), u \in \mathcal{N}(v) \}))
\end{equation}
where $h_v^k$ is the node feature of center node $v$  at $k$th layer, $e_{uv}$ is the edge feature between $v$ and its neighbor node $u$. $\mathcal{N}(v)$ denotes the neighborhood of node $v$, usually 1-hop neighbors.  Aggregating function $\mathbf{A(\cdot)}$ aggregates node features over neighborhood of $v$, while updating function $\mathbf{C(\cdot)}$ integrates features both from center node $v$ and its neighboring  features aggregated by  $\mathbf{A(\cdot)}$. Various mathematical operations have been adopted as $\mathbf{A(\cdot)}$ and $\mathbf{C(\cdot)}$. For instances, in work \cite{hamilton2017inductive}, mean-pooling and attention mechanisms are used as  $ \mathbf{A(\cdot)}$, whereas GRU \cite{li2015gated}, concatenation \cite{hamilton2017inductive} and summation are usually applied as $\mathbf{C(\cdot)}$.

For obtaining whole graph representation, a pooling function, namely read-out, is used at the last $K$th layer:
\begin{equation}
\label{eq:rdout}
h_G = \mathbf{R}(\{ h_v^K | v \in G\})
\end{equation}
$h_G$ means the whole representation of input graph $G$. $\mathbf{R}(\cdot)$ represents the read-out function.

Spectral based GNN methods require Fourier transform on graph,  meaning that all the  input graph samples should be static (i.e., fixed topological structure) \cite{kipf2016semi}. In contrast, spatial based GNN methods have no such restriction, and  are able to extract features on graphs with varied structures. Hence, spatial based GNNs are suitable for  the 1-hop subgraph pairs from bipartite networks in our work.

\subsection{Molecule  Representation}
\label{sec:molrepresent}
For the  molecule chemical representation, current GNN methods  \cite{cherkasov2014qsar, matveieva2021benchmarks, tang2020self} usually process 2D graph as  description of natural chemical graph, in  which nodes  represent  atoms integrating different chemical attributes,  and edges represent  bonds  connecting atoms to one another. There are mainly three advantages of using 2D graph description: (1) graph  preserves clear and stable information of chemical structure, (2) it   represents invariant molecule regardless of entry position in line notation (e.g., SMILES \cite{weininger1988smiles}), (3) it can be easily computed and optimized by GNN methods.  In our chemical formulation, we take similar 2D graph representation, meanwhile we adopt bidirectional graph where the bond connection from atom A to atom B is the same as the bond connection from atom B and atom A. Moreover, 7 atomic chemical attributes, listed in Figure \ref{fig:framework} (II), are considered as node initial features of input graph. Our GNN method learns and aggregates these attributes to be proper  molecular features. For data preprocessing,  we convert SMILES sequence into  2D graph formula, and node attributes in Figure \ref{fig:framework}:II can be used for distinguishing those compounds with same molecular structures. Hence, each molecule representation in our method is unique.

\section{Molecule-Target Interaction Network plus (MTINet+)}
\label{sec:gplpintro}
In this part, we introduce our end-to-end pseudo-siamese GNN method , namely MTINet+, for predicting potential/missing links in biomedical network. For a given bipartite biomedical network $\mathcal{G}$, we constructed the observed interaction matrix $\mathcal{M} \subset \mathbb{R}^{M \times N}$, where $M$ is the molecule number and $N$ is the target number, each row index ($m \in \mathbb{Z}^M$) and column index ($t \in \mathbb{Z}^N$)   denote the sequential identical numbers of molecules ($\Omega^M$) and targets  ($\Gamma^N$) respectively, and each entry $y_{m,t}$ in $\mathcal{M}$ represents whole possible interaction of $(m, t)$ pair. In our case, $y_{m,t} \in [0, 1, x]$, $1$ denotes observed positive label meaning active interaction between $m$ and $t$, whereas $0$ is the observed negative label meaning the inactive interaction. Here $x$ means the unobserved (missing) interaction waiting for prediction.

\subsection{1-hop Subgraphs Pair Construction}
\label{sec:1-hop}
In the bipartite network $\mathcal{G}$, for each molecule $m$ and target $t$  we respectively exact their 1-hop subgraphs $G^M_{m \to t}$ and  $G^N_{t \to m}$, in which the edges $E^M_{m \to t}$ and $E^N_{t \to m}$ mean the links (active interactions) of neighbouring target nodes ($\hat{t} \in G^M_{m \to t} $) around center node $m$, and neighbouring molecule nodes ($\hat{m} \in G^N_{t \to m} $) around center node $t$ respectively.  Then  for each $(m,t)$ , we pair $G^M_{m \to t}$ and $G^N_{t \to m}$ as a regrouped graph $\hat{G}_{m,t}^{M \times N}$, and label $\hat{G}_{m,t}^{M \times N}$ with $(m,t)$ corresponding $y_{m,t}$ in $\mathcal{M}$. We feed the paired subgraph dataset  to the GNN model for training or prediction. Note that after pairing the subgraphs, if $y_{m,t}$ is positive, the $(m,t)$ edge and corresponding nodes should be removed from  both $G^M_{m \to t}$ and $G^N_{t \to m}$. It is to make sure that GNN model cannot see any link for prediction in the training subgraphs pair.  In Figure \ref{fig:framework} (I), steps $(a)$, $(b)$,  $(c)$ and $(f)$ show  MTINet+ procedures  that convert the bipartite network into graph structure data for one  MTINet+ GNN branch to learn  the network topological  representation.  

\subsection{Molecule Structural and Chemical Representation}
\label{sec:molchemrepresent}
For each molecule $m$, we utilize Python RDkit \cite{landrum2013rdkit} tool kit to process molecule SMILES as input data format, and convert the data into bidirectional graph $S_m^M$ based on Deepchem and Chemprop \cite{ramsundar2019deep, yang2019analyzing} processing ways. The graph mainly consists of index lists of nodes and edges. For each node, each attribute in Figure \ref{fig:framework} (II) is converted into identical number, and all 7 numeric attributes are  combined as initial node feature vector $f_{init}$. The  node feature $f_{init}$ is then delivered to node embedding layer of our framework for learning optimal attribute combination, from which we can  obtain an aligned node feature vector $f_v$.   In Figure \ref{fig:framework}(I), steps $(e)$ and $(g)$ show  MTINet+ procedures  that convert  the molecule SMILES into bidirectional graph structure data for the other  MTINet+ GNN branch to learn  the molecule structural and chemical representation.

\subsection{ Pseudo-Siamese Graph Neural Networks Architecture}
\label{sec:gnnmodel}
As mentioned above, the MTINet+ backbone consists of two Graph Neural Networks branches, they have the same algorithm structure, a variant of GIN \cite{xu2018powerful},  which belongs to spatial based graph neural networks. As the two GNN branches do not share weights, the MTINet+ backbone forms a pseudo-siamese GNN architecture. Such architecture has demonstrated good performance on learning cross-domain features \cite{gong2023cross}. 

Each GNN branch aggregates features by taking both  summation and maxima of neighbouring features, and thus enhances the message propagation from shallow layers to deep layers.  Respectively, for each node $v$ in graph $\hat{G}_{m,t}^{M \times N}$ or graph $S_m^M$, its features $h_v^k$ out of $k$th layer can be uniformly written as:
\begin{equation}
\label{eq:hagnet}
h_v^k = \phi(\mathbf{concat}(h_v^{k-1}, (\sum_{u \in \mathcal{N}(v)}{h_u^{k-1}} + \max_{u \in \mathcal{N}(v)}{h_u^{k-1}} )))
\end{equation}
where $\phi(\cdot)$ is the MLP function, and $\mathbf{concat(\cdot)}$ concatenates  features of node $v$ from $(k-1)$th layer and  $k$th layer, $u$ denotes a neighbouring node in node $v$'s neighbourhood $\mathcal{N}(v)$. The output node feature $h_v^k$ are concatenated with node features  from all the previous layers, i.e., $H_v^k = \mathbf{concat}(h_v^k, h_v^{k-1}, ...,  h_v^1 )$. Then we use mean pooling \emph{read-out} function to obtain final representations $H_{\hat{G}_{m,t}}$ of graph $\hat{G}_{m,t}^{M \times N}$, and $H_{S_m}$ of graph $S_m^M$ respectively. As  mentioned in \ref{sec:1-hop}, $\hat{G}_{m,t}^{M \times N}$ consists of  $G^M_{m \to t}$ and $G^N_{t \to m}$ pairs. Thus $H_{\hat{G}_{m,t}}$  is learnt as one representation for both $G^M_{m \to t}$ and $G^N_{t \to m}$.

Finally, the  representations $H_{\hat{G}_{m,t}}$  and  $H_{S_m}$ are concatenated and fed to a 3-layer MLP $\Phi(\cdot)$  functional head for learning to predict interaction (link) probability $\hat{y}_{m,t}$:
\begin{equation}
\label{eq:mlphead}
\hat{y}_{m,t} = \Phi(\mathbf{concat}(H_{\hat{G}_{m,t}},H_{S_m}))
\end{equation}

\subsection{Loss Function and Optimization}
Since the link prediction in our case  is  a binary classification task, i.e. predicting active or inactive interaction, we adopt cross-entropy \cite{murphy2012machine} as our loss function, which can be described as:
\begin{equation}
\label{eq:loss}
\mathcal{L}_{m,t} = -(y_{m,t}\log{\hat{y}_{m,t}} + (1- y_{m,t})\log{(1-\hat{y}_{m,t})})
\end{equation}
where $y_{m,t}$ is the observed interaction, and $\hat{y}_{m,t}$ is the predicted link probability.
For optimization on model parameters, we tried SGD \cite{kiwiel2001convergence}, Adam  \cite{kingma2014adam}  as optimizers, and find that Adam gives the best optimized  model with highest evaluation performance and most stable convergence during training and testing.

\section{Dataset and Experiments}
\label{sec:data&exp}
We conducted experiments on three heterogeneous biomedical networks: Drug-Target Interaction (DTI) dataset from DTINet \cite{luo2017network}, National Toxicology Program (NIH) Tox21 challenge dataset \cite{mayr2016deeptox}, and GHDDI constructed Compound-Virus Inhibition (CVI56) Dataset. We trained our MTNet+ models on each dataset respectively, and tuned each model's parameters with different protocols. In details, we took a 80-20 split on datasets of DTI and CVI56 respectively, where 80\% of dataset was used for training and the remaining 20\% was used for testing. For model training on NIH Tox21 dataset, we followed the 5-cross validation protocol from the baseline work \cite{feinberg2018potentialnet}. We took a random shuffle on each dataset for ensuring each input data sample give an independent change on the model in each training batch. Also, we resmapled the training positive and negative  data samples, balancing their number ratio to 1:1, it avoids the model's skewness towards the class with majority number during training. For the hyperparameters configuration, we set 5 GNN layers for each backbone branch, and selected Adam optimizer with learning rate $\eta = 0.001$, $\xi = 10^{-16}$ , $(\beta_0,  \beta_1) = (0.9, 0.999)$ and weight decay $\lambda = 0.0001$. The training batch size and epoch number were fixed to 256 and 1,000 for  all datasets. This work is publicly available on GitHub: https://github.com/GHDDI-AILab/MTINetplus . 

\subsection{DTI Dataset from DTINet}
\begin{figure}
	\centering
	\includegraphics[scale=.4]{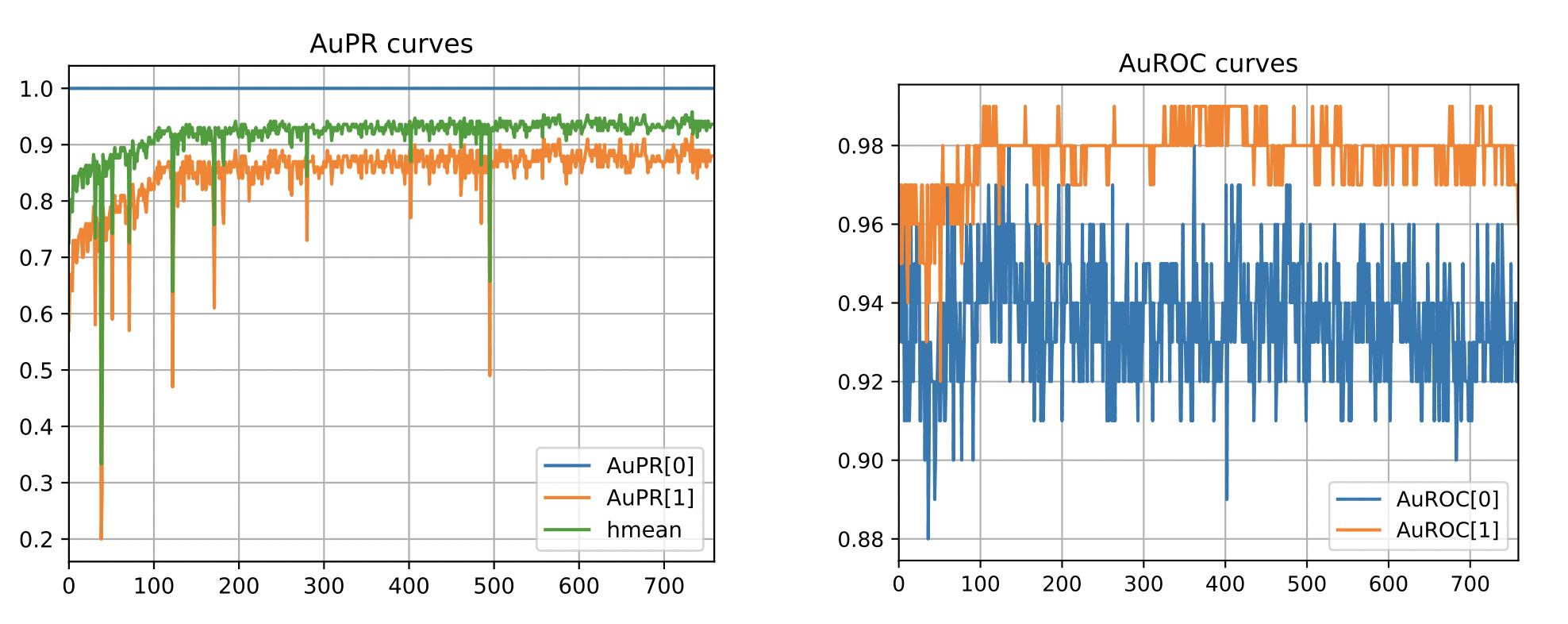}
	\caption{Traing procedure of MTINet+ model on DTI  dataset.}
	\label{fig:trainingDTI}
\end{figure}

We chose the heterogeneous network constructed in the work \cite{luo2017network}. The network includes 12,015 nodes and 1,895,445 edges in total, for predicting missing drug-target interactions. It incorporates 4 types of nodes (i.e., drugs, proteins, diseases and side-effects) and 6 types of edges (i.e., drug-protein interactions, drug-drug interactions, drug-disease associations, drug-side- effect associations, protein-disease associations and protein-protein interactions). Besides other well known link prediction methods, we also introduced our previously designed GPLP \cite{guo2021heterogeneous} method as comparison. GPLP is a topological information based GNN framework, and independent on node's side information. Figure \ref{fig:trainingDTI} shows the convergence of AUROC and AUPR curves during the training procedure of MTINet+ model. Compared with all the counterparts, our MTINet+ model constantly outperforms in terms of AUROC and AUPR, reaching up to 96\% and 93\% respectively. The AUROC and AUPR we computed are the harmonic average between positive and negative performances. Our MTINet+ model performs nearly 5\% higher than DTINet and 2\% higher  than GPLP in terms of AUROC, and and 1\% higher than DTINet and 2\% higher  than GPLP  in terms of AUPR. MTINet+ also beats other Random Walk based methods (e.g., HNM\cite{wang2014drug}), see details in Table \ref{tab:DTIcomp}.

\begin{table}
	\caption{Different methods comparison on DTI dataset. Our MTINet+ outperforms other state-of-the-art methods for DTI prediction.}
	\centering
	\begin{tabular}{cccccccc}
		\toprule
		Methods     & BLMNII \cite{mei2013drug}     & NetLapRLS \cite{xia2010semi}	&HNM \cite{wang2014drug}		&CMF \cite{zheng2013collaborative} 		&DTINet \cite{luo2017network} 		&GPLP \cite{guo2021heterogeneous} &MTINet+(ours) \\
		\midrule
		AUROC		 & 0.67  			& 0.83			&0.86		&0.87		&0.91			&0.95 &\textbf{0.96} \\
		AUPR		 & 0.74				& 0.88    		&0.88		&0.86 		&0.93			&0.92  &\textbf{0.94}	\\
		\bottomrule
	\end{tabular}
	\label{tab:DTIcomp}
\end{table}

\subsection{NIH Tox21 Challenge Dataset}
\begin{figure}
	\centering
	\includegraphics[scale=.4]{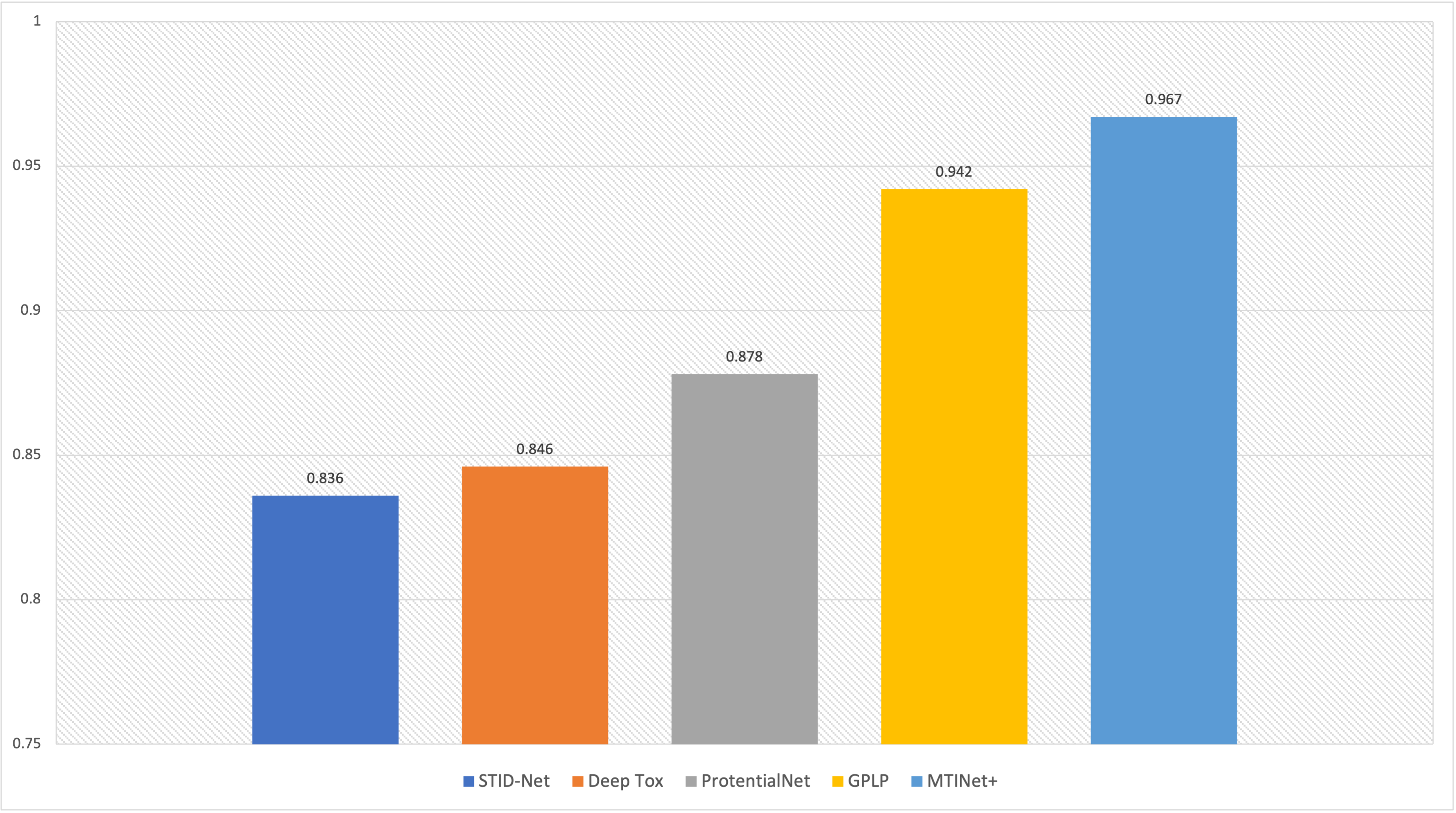}
	\caption{Different methods AUROC comparison on NIH Tox21 dataset. MTINet+ outperforms other state-of-the-art methods for toxicity prediction.}
	\label{fig:tox21comp}
\end{figure}

For NIH Tox21 Challenge Dataset \cite{mayr2016deeptox}, a dataset with 12,707 chemical compounds, which consisted of a training dataset of 11,764, a leaderboard set of 296, and a test set of 647 compounds. For the training dataset, the chemical structures and assay measurements for 12 different toxic effects were fully available at the beginning of the challenge, so were the chemical structures of the leaderboard set.To fulfil the challenge, the common methods, no matter what kind: descriptor based \cite{cao2012kernel, darnag2010support, sagardia2013new} or deep learning methods \cite{mayr2016deeptox, feinberg2018potentialnet}, focus on extracting effective representations of chemical compounds under certain toxicity task. In contrast, our prospective takes an insight  into the interactions (toxic or non-toxic) between compound (molecule) and toxicity task (target), regardless of compound chemical structures and task properties. In the experimental results, we find that our MTINet+ model significantly outperforms against the  chemical structure based deep models (e.g., DeepTox \cite{mayr2016deeptox}, ProtentialNet \cite{feinberg2018potentialnet}), achieving AUROC of 96.7\% and AUPR of 95.8\%. In addition to GPLP, we previously designed a chemical structure based GNN model for multi-task prediction, namely STID-Net, which can compete with other  chemical structure based deep methods. Hence, we also added  performance  of STID-Net into comparison (see Figure \ref{fig:tox21comp}).

\subsection{CVI56 Dataset}
During COVID-19 pandemic broke-out in the year 2020, we constructed an antiviral compound-phenotype network, which collected  9,196 drugs (molecule), 56 virus families (target) and their observed 12,196. interactions. The training dataset contains 7,234 active (IC50<=1uM) and 4,962 inactive (IC50>1uM) interactions.  For this newly constructed dataset, we conducted the experiments with our MTNet+ model and GPLP as benchmark. MTINet+ model achieved 93.1\% in AUROC, while GPLP performs a little better, got 94.6\% in AUROC. The CVI56 Dataset is publicly available on GitHub: https://github.com/GHDDI-AILab/MTINetplus.

\section{Discussion and Future Work}
\label{sec:dis&future}
\subsection{Robustness Verification and Comparison}
\begin{figure}
	\centering
	\includegraphics[scale=.25]{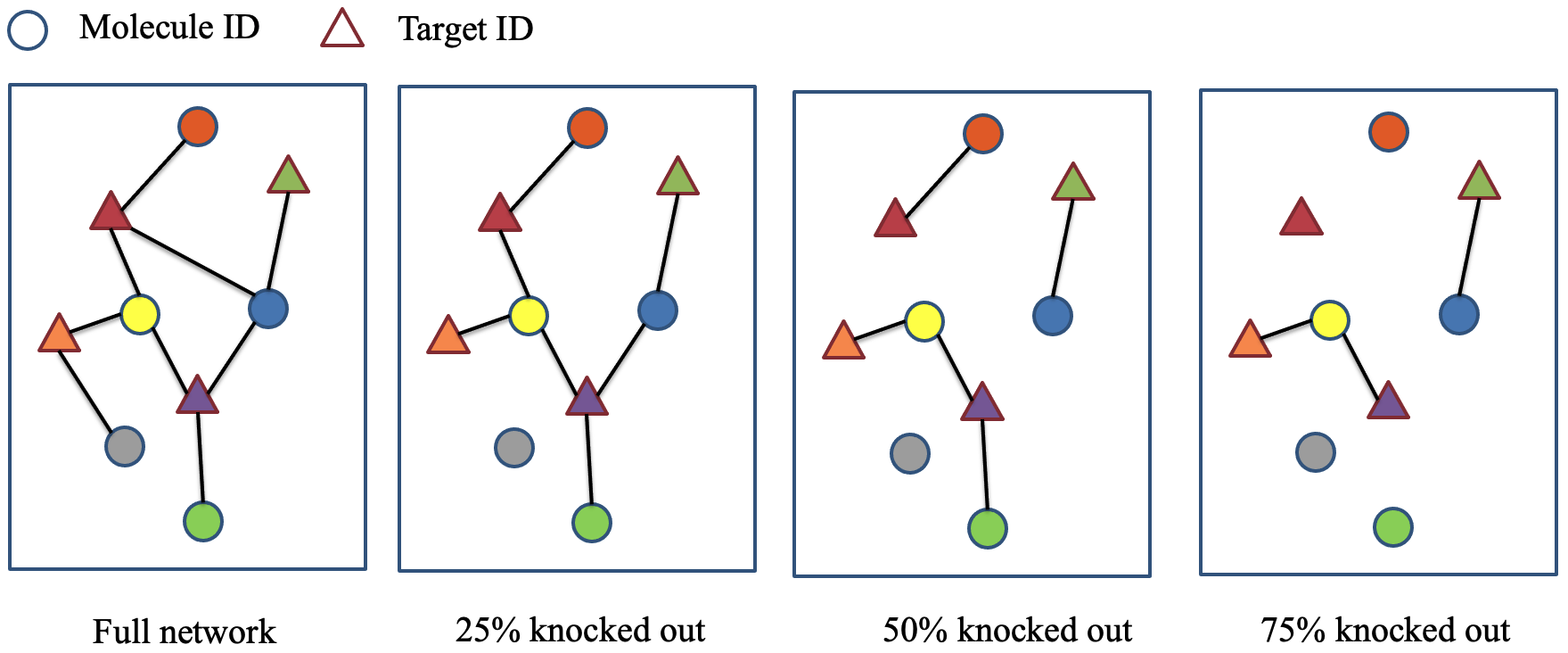}
	\caption{Different extents of knocking out links in the network to  simulate the partially observed networks in the real world scenarios.}
	\label{fig:randomknock}
\end{figure}

As in  practical biomedical problems, usually the observed links in the network are rather limited and incomplete, and thus it causes a sparse observed matrix \cite{luo2017network}. It implies that the observed links only cover a very small portion of real biomedical network  (i.e., the observed matrix is of very low-rank ). This fact results in a problem that \emph{local graph pattern} leant from training data cannot fully reveal the reality. However, besides the network information, MTINet+ also takes the molecule structural/chemical information for predicting interaction link. Therefore, MTINet+ should overcome the sparse biomedical network issue.  Unfortunately, the network that covers the \emph{whole} connections  can be never obtained  in real world as our experimental dataset. To analyse our MTINet+ model's robustness against real scenarios, we assume that original network we have  covers the  \emph{whole} connections, and we randomly knock-out connections in the network to simulate the partially observed network in reality (see Figure \ref{fig:randomknock}). Then, we trained MTINet+ model on these partial network datasets. 

\begin{figure}
	\centering
	\includegraphics[scale=.6]{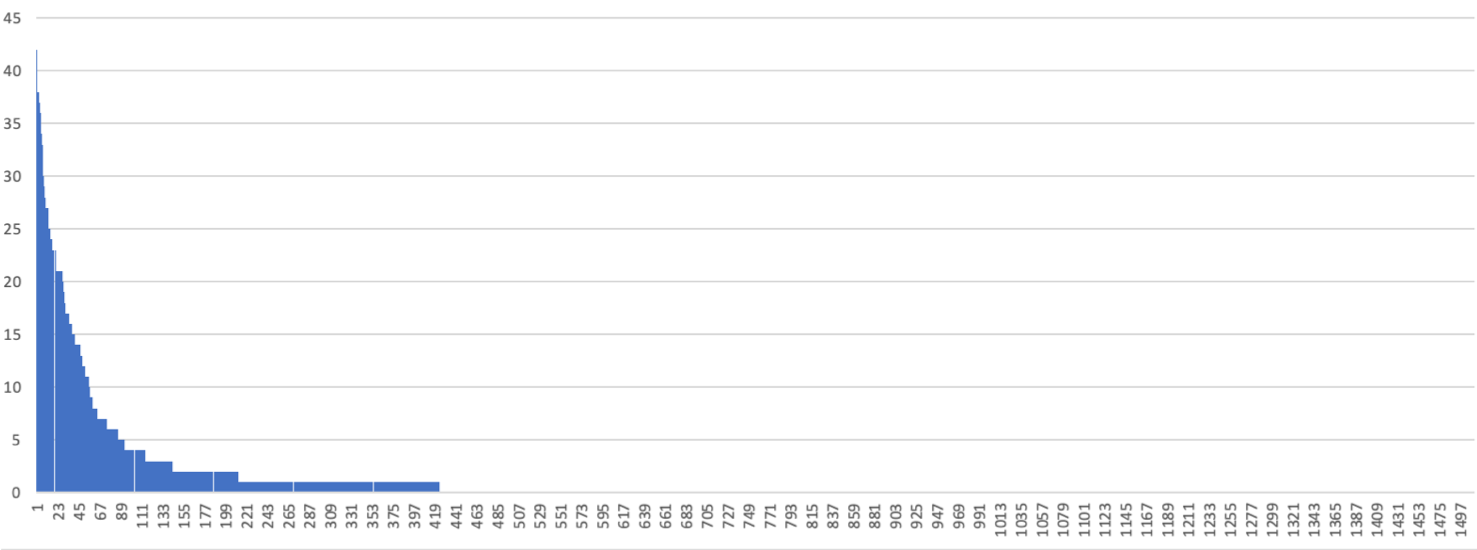}
	\caption{molecule-target subgraph  $G^M_{m \to t}$ degree distribution on DTI dataset. A long-tail distribution.}
	\label{fig:degreedistrbdti}
\end{figure}

To implement the knocking-out protocol, we firstly convert the whole network into subgraph pairs  $G^M_{m \to t}$ and $G^N_{t \to m}$, and then randomly knock-out the edges of $G^M_{m \to t}$ and $G^N_{t \to m}$ respectively, obtaining ${G'}^M_{m \to t}$ and ${G'}^N_{t \to m}$. However, in the biomedical dataset, the degree distribution of all the $G^M_{m \to t}$ (or $G^N_{t \to m}$) usually appears as long-tail. For instance, Figure \ref{fig:degreedistrbdti} shows $G^M_{m \to t}$ degree  distribution of DTI dataset, where horizontal axis denotes the molecule (drug) ID and vertical axis means the degree of $G^M_{m \to t}$. As we can see, most of molecules have 1 or even 0 interaction with any target. In order to equally knock each subgraph, we invented  a knocking-out method by random sampling a knocking-out portion according to the mixture distributions of  $G^M_{m \to t}$ edges, which is conducted as:

\begin{equation}
\label{eq:mixtureditb}
\begin{split}
& {E'}^M_{m \to t} = \rho_\delta \oslash E^M_{m \to t}  \\
&  \rho_\delta  =  \frac{C_\Delta^\delta}{2^\Delta-1},  \;  \delta \in \Delta  
\end{split}
\end{equation}
where ${E'}^M_{m \to t}$ denotes the edges after knocking-out on original edges $E^M_{m \to t}$ of $G^M_{, \to t}$. $ \rho_\delta$ is a probabilistic function which determines the knocking-out portion of $E^M_{m \to t}$ , sign $\oslash$ means the random edge removal according to $ \rho_\delta$,  $\Delta$ is the degree of $G^M_{m \to t}$, while $C_\Delta^\delta$ means the $\delta$-combinations over $\Delta$. The knocking-out operation on $G^N_{t \to m}$ follows the same way as  $G^M_{m \to t}$ .

\begin{figure}
	\centering
	\includegraphics[scale=.4]{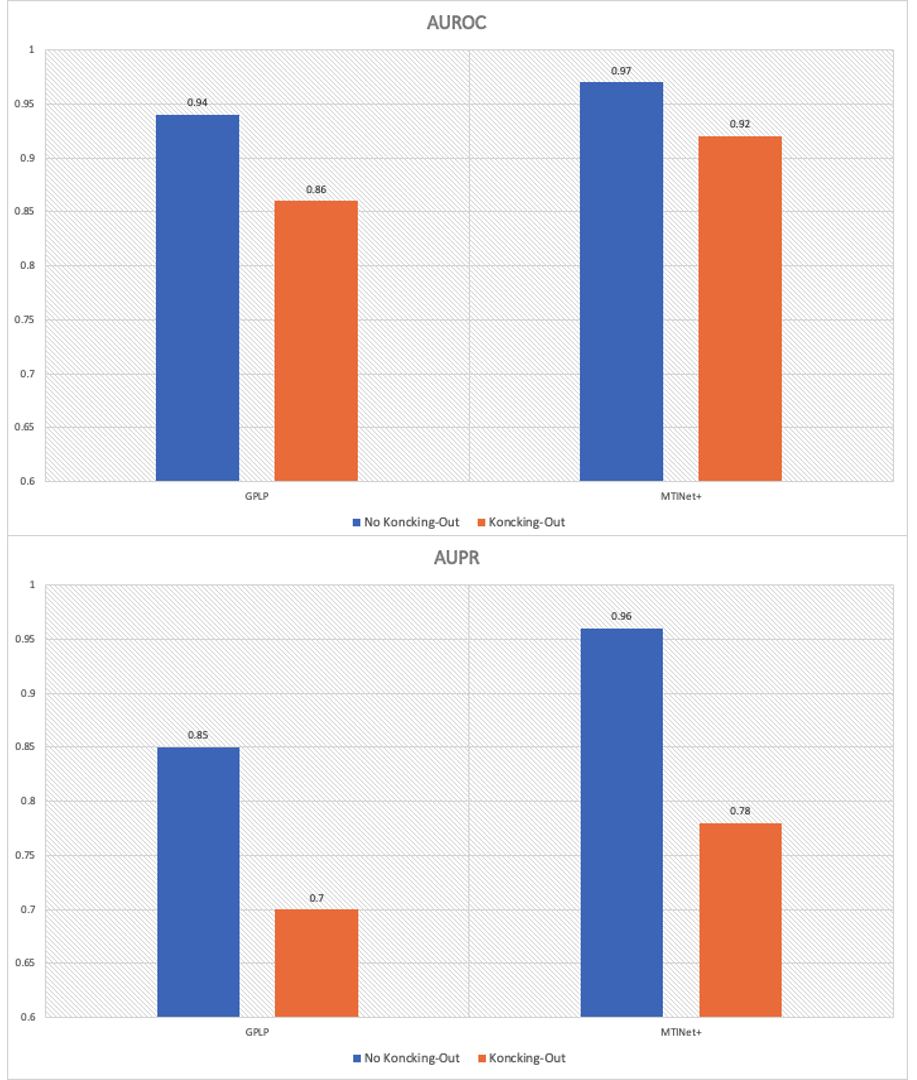}
	\caption{The robustness comparison of MTINet+ and GPLP with the knocking-out protocol on NIH Tox21 dataset.}
	\label{fig:knockouttox21}
\end{figure}

We compared the perfomance of MTINet+ and GPLP (only network information based GNN method) with this knocking-out protocol on the NIH Tox21 dataset. As was expected, the performances of both MTINet+ and GPLP models trained on knocked-out datasets declined w.r.t. AUROC and AUPR, since the topological information was constantly lost with varying degrees during training. Specifically,  the MTINet+ model got the stable AUROC of 92\% ( 5\% decreased) and AUPR of 78\%( droping 18\%), whereas the GPLP model converged at the AUROC of 86\% ( 8\% decreased) and AUPR of 70\%( droping 15\%). Figure \ref{fig:knockouttox21} shows the robustness comparison of MTINet+ and GPLP.  As we can see, even though the performances of  both MTINet+ and GPLP models  dropped on partially knocked-out networks,  MTINet+  still gives better results than GPLP, which implies that the molecule information contributes to the resistance of limited network information. Thus, MTINet+ demonstrates the robustness in  real world scenarios.

\subsection{Conclusions and Future Work}
In this work, we have introduced our pseudo-siamese Graph Neural Network framework, namely MTINet+, for  predicting potential/missing links in biomedical networks. MTINet+ learns both biomedical network topological  and molecule structural/chemical information  as  representations to predict potential interaction of given molecule and target pair.  Our method has demonstrated an out-performance compared to the counterparts. In addition, since MTINet+ incorporates both molecule and network information, it naturally solves the cold start problem and shows strong robustness  against different sparse biomedical networks. However, it is also important to consider target information (protein sequence, pocket structure, etc.)  for providing more robust and accurate link prediction. Hence, our key work in future is to incorporate this three heterogeneous information (molecule, target and their biomedical network) and build a competent deep learning framework.  Finally, MTINet+ shows a novel perspective for revealing the underlying interaction mechanisms of complex biomedical system. 

\bibliographystyle{unsrtnat}
\bibliography{references}  
\end{document}